\documentclass[letterpaper, 10 pt, conference]{ieeeconf}  % Comment this line out if you need a4paper

\IEEEoverridecommandlockouts                              % This command is only needed if 
\usepackage{times}
\usepackage{helvet}
\usepackage{courier}
\usepackage{multicol}
\usepackage{comment}
\usepackage{graphicx}
\usepackage{amsmath}
\usepackage{stmaryrd}
\usepackage{paralist}
\usepackage{url}
\usepackage{hyperref}
\usepackage{cleveref}
\usepackage[outdir=./images/]{epstopdf}
\usepackage[linesnumbered,ruled]{algorithm2e}
\newcommand{\U}{\mathcal{U}\xspace}
\newcommand\M{{\cal M}}        % caligraphic Mß

\newcommand{\set}[1]{\{#1 \}}
\newcommand{\?}{\raisebox{.5pt}{\textcircled{\raisebox{-.9pt} {{\small
          ?}}}}}

\newcommand{\citep}[1]{\cite{#1}}

\newtheorem{definition}{Definition}

\newtheorem{theorem}{Theorem}
\usepackage{xcolor}
\usepackage{etoolbox}
\usepackage{caption}
\usepackage{subcaption}
\providetoggle{full}

\crefformat{section}{\S #2#1#3} 
\crefformat{subsection}{\S #2#1#3}
\crefformat{subsubsection}{\S #2#1#3}

\title{Anytime Integrated Task and Motion Policies for Stochastic Environments}

\author{Naman Shah, Deepak Kala Vasudevan, Kislay Kumar, Pranav Kamojjhala,  and Siddharth Srivastava \\ 
School of Computing, Informatics, and Decision Systems Engineering,\\
Arizona State University, Tempe, AZ, USA\\
\{namanshah,  dkalavas, kkumar28, pkamojjh, siddharths\}@asu.edu
}% <-this % stops a space
\begin{document}

\maketitle
\thispagestyle{empty}
\pagestyle{empty}

\begin{abstract}
  In order to solve complex, long-horizon tasks, intelligent robots
  need to carry out high-level, abstract planning and reasoning in
  conjunction with motion planning. However, abstract models are
  typically lossy and plans or policies computed using them can be
  unexecutable. These problems are exacerbated in 
  stochastic situations where the robot needs to reason about, and
  plan for multiple  contingencies.
    We present a new approach for integrated task and
  motion planning in stochastic settings. In contrast to prior work in this
  direction, we show that our approach can effectively compute
  integrated task and motion policies whose branching structures
  encoding agent behaviors handling multiple execution-time contingencies. We
  prove that our algorithm is probabilistically complete and can
  compute feasible solution policies in an anytime fashion so that the
  probability of encountering an unresolved contingency decreases over
  time. Empirical results on a set of challenging  problems show the
  utility and scope of our methods.
\end{abstract}
\settoggle{full}{false}
\section{Introduction}

\begin{figure}[t!]
    \centering
    \includegraphics[height=1.5in]{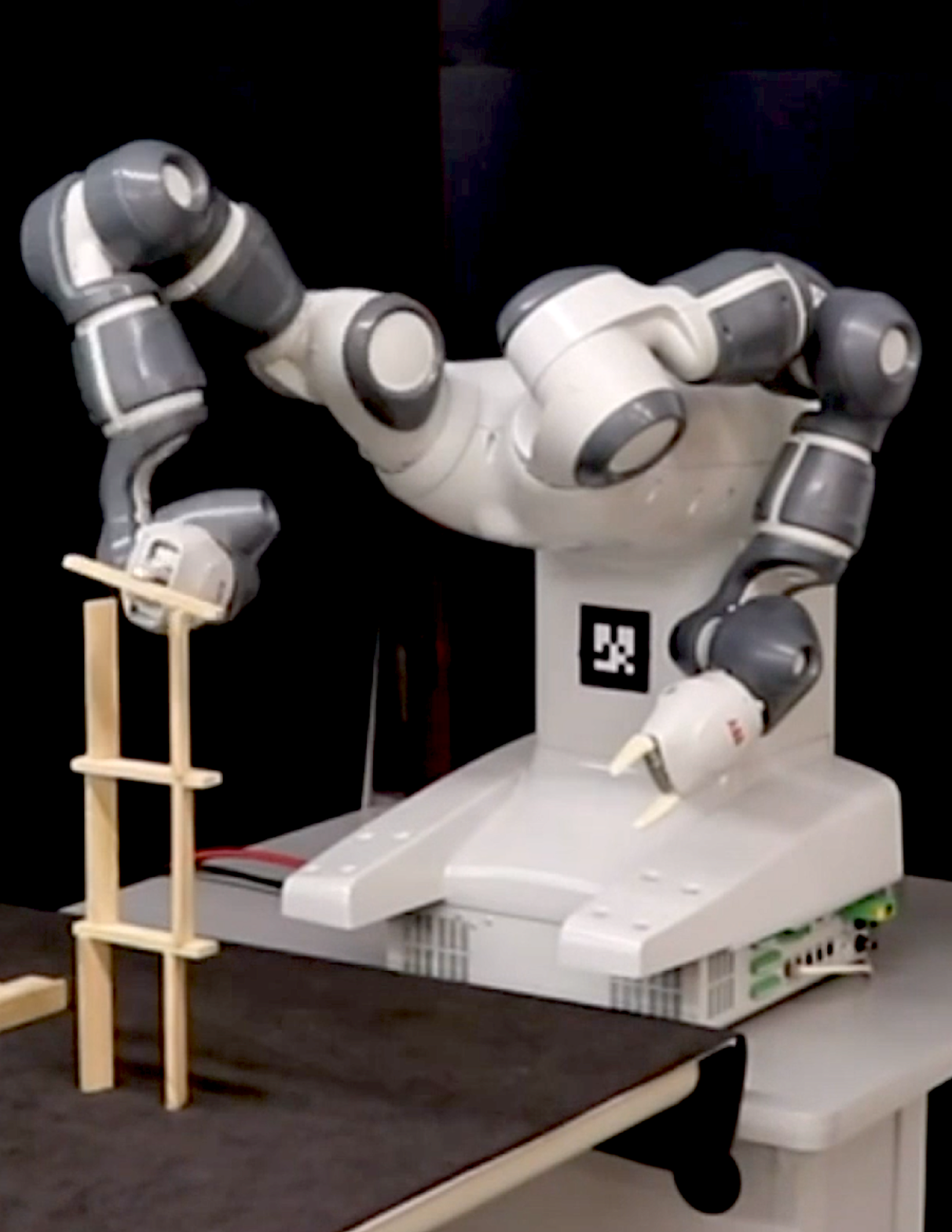}\hspace{1cm}
    \includegraphics[height=1.5in]{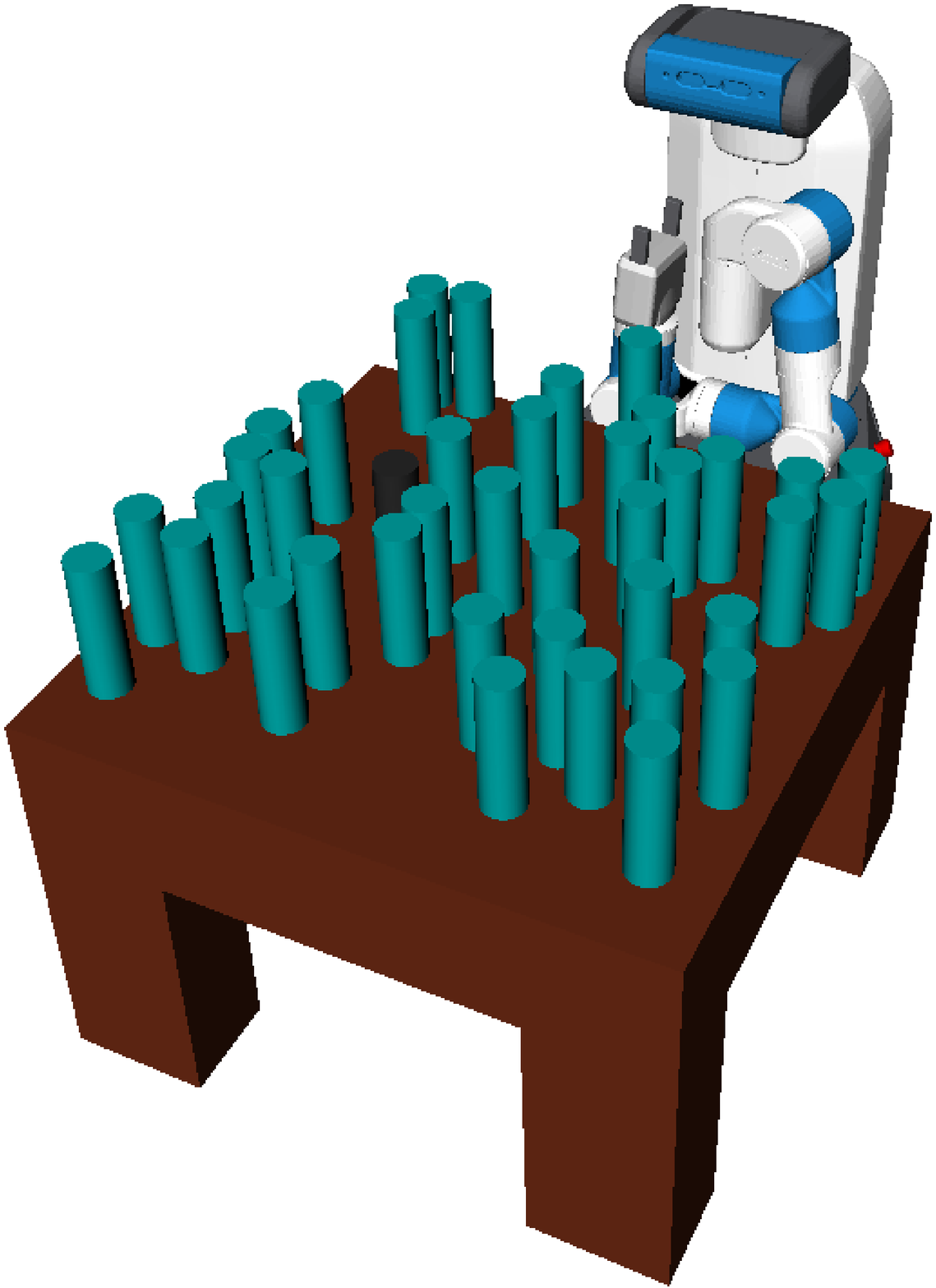}
    \caption{Left: YuMi robot uses the algorithm developed in this paper to build a $3\pi$ structure using Keva
      planks despite stochasticity in their initial locations. Right:
      A stochastic variant of the cluttered table domain where robot
      has to pick up the black can, but pickups may fail.}
    \label{fig:domainFig}
\end{figure}

Recent years have witnessed immense progress in research on integrated
task and motion planning
\cite{ffrob,dantam2018incremental,cashmore2015rosplan,pddlstream,chitnis16_guidedTMP,srivastava14_tmp}. Research
in this direction provides several approaches for solving
deterministic, fully observable task and motion planning
problems. However, the problem of integrated task and motion planning
under uncertainty has been under-investigated.  We consider integrated
task and motion planning problems where the robot's actions and its
environment are stochastic.  This problem is more difficult
computationally because sequential plans are no longer sufficient;
solutions take the form of \emph{policies} that prescribe an action
for every state that the robot may encounter during execution.  For
instance, consider the problem where a robot needs to pick up a can
(black) from a cluttered table (Fig. \ref{fig:domainFig}). To achieve
this objective, the robot needs to consider multiple contingencies,
e.g., what if the can slips? What if it tumbles and rolls off when it
is placed?

This example is representative of many real-world problems such as
diffusing IEDs, operating live machinery, or assisting emergency
response personnel. Safe robot execution in such situations requires
pre-computation of truly feasible policies so as to reduce the need
for time-consuming and error-prone on-the-fly replanning.
A na\"ive approach for solving such problems would be to first compute
a high-level policy using an abstract model of the problem (e.g., a
model written in a language such as PPDDL or
RDDL\cite{sanner10_rddl}), and to then refine each ``branch'' of the
solution policy with motion plans. Such approaches fail because
abstract models are lossy and policies computed using them might not
have any feasible motion planning refinements
\cite{cambon09_asymov,kaelbling11_hierarchical,srivastava14_tmp}. Furthermore,
as the planning horizon increases, computing complete task and motion
policies becomes intractable as it requires the computation of
exponentially many task and motion plans, one for each branch of the policy.

We present a novel \emph{anytime} framework for computing integrated
task and motion policies. Our approach continually improves the
quality of solution policies while ensuring that the versions computed
earlier can resolve situations that are more likely to be encountered
during execution. It also provides a running estimate of the
probability mass of likely executions covered in the current
policy. This estimate can be used to start execution based on the
level of risk acceptable in a given application, allowing one to
trade-off precomputation time for on-the-fly invocation of our planner if an
unhandled situation is encountered.  Our approach generalizes methods
for computing solutions for most-likely outcomes during
execution~\cite{hadfield15_modular,determinization} to the problem of
integrated task and motion planning by drawing upon approaches for
anytime computation in AI
planning~\cite{dean88_anytime,zilberstein93_anytime,dean95_anytime}. Our
experiments indicate the probability of encountering an unresolved
contingency drops exponentially as the algorithm proceeds.  The
resulting approach is the first \emph{probabilistically complete}
algorithm for computing integrated task and motion policies in
stochastic environments using a powerful relational representation for
specifying input problems. % The use of relational representations
% allows us to easily express problems involving object manipulation,
% which would be cumbersome if not infeasible in propositional
% representations.
Our approach uses arbitrary stochastic shortest path (SSP) planners
and motion planners.  This structure allows it to scale automatically
with improvements in either of these active areas of planning
research.

% We utilise the Markov decision processes (MDP) for modeling the
% problems that need to be solved by the robot. The MDP framework can
% express sequential decision making processes with stochastic action
% outcomes. Several approaches have been designed for solving MDPs. The
% scalability of these approaches relies upon the assumption of the
% bounded number of possible actions and the representation of the state
% accurately with discrete variables. However, both of these assumptions
% fail in the problem stated above. Recent work on integrated task and
% motion planning
% (e.g., \cite{kaelbling11_hierarchical,srivastava2014combined,dantam16_incremental})
% shows that the hierarchical approaches can be used effectively for
% such problems.

We begin with a presentation of the background definitions
(\ref{sec:background}) and our formal framework
(\ref{sec:formal}). (\ref{sec:algo}) describes our overall algorithmic
approach, followed by a description of empirical results using the
Fetch and YuMi robot platforms (\ref{sec:results}), and a discussion of
prior related work (\ref{sec:related}).

\section{Background}
\label{sec:background}

A \emph{motion planning problem} is a tuple
$\langle C, f, p_0, p_t \rangle$, where $C$ is the space of possible
configurations or poses of a robot, $f(p)$ is a boolean function which
determines whether the robot at config $p \in C$ is in collision with any object or not, and $p_0, p_t \in C$ are the initial and final configs. A trajectory is
a sequence of configurations. A collision-free motion plan
for a motion planning problem is a trajectory in $C$ from $p_0$ to
$p_t$ such that $f$ is false for any pose in the trajectory.

% A \emph{Markov decision process} (MDP) is defined as a tuple $(S,A,T,R,\gamma)$ where \begin{inparaitem}
%     \item[] $S$ is a set of states;
%     \item[] $A$ is a set of possible of actions;
%     \item[] $T$ is a transition function where $T(s,a,s')=P(s'|s,a)$ for $s,s' \in S, a \in A $;
%     \item[] $R(s,a,s')$ is a reward function for $s,s' \in S, a \in A $;
%     \item[] $\gamma \in [0,1]$ is the discount factor.
% \end{inparaitem}
% A solution to an MDP is a \emph{policy}, $\pi: S\rightarrow A$, which
% maps each state to an action.

\emph{Stochastic shortest path (SSP)} problems are a subclass of Markov
decision processes (MDPs) that have absorbing states, the discounting
factor $\gamma = 1$ and a finite horizon~\cite{bertsekas91_ssp}. An
SSP can be defined as a tuple $\langle S,A,T,\mathcal{C},\gamma=1,H,s_{0},G\rangle$
where $S$ is a set of states; $A$ is a set of actions; $\forall s,s'\in S,\quad a\in A\quad T(s,a,s')=P(s'|s,a)$;
\begin{inparaitem}
    \item[] $\mathcal{C}(s,a)$ is the cost for action $a \in A$ in a state $s \in S$;
    \item[] $H$ is the length of  the horizon;
    \item[] $s_{0}$ is an initial state;
    \item[] $G$ is the set of absorbing or goal states.
\end{inparaitem} A solution to an SSP is a policy $\pi$ of the form
$\pi:S\times \{1,\dots,H\} \rightarrow A$ that maps all
the states and time steps at which they are encountered to an
action. The optimal policy $\pi^{*} $ is a policy that reaches the
goal state with the least expected cumulative cost. SSP
policies need not be stationary because the horizon is finite. Dynamic
programming algorithms such as value iteration or policy iteration can be
used to compute these policies. Value iteration for finite horizon SSPs can be defined as:
\begin{align*}
    V^{0}(s) &= 0\\
    V^{i}_{t}(s) &= min_{a}\sum_{s'}T(s,a,s')\left[ \mathcal{C}(s,a) + V^{i-1}_{t+1}(s')\right]\\
  \pi^{i}_{t}(s) &= argmin_{a}\sum_{s'}T(s,a,s')\left[ \mathcal{C}(s,a) + V^{i-1}_{t+1}(s') \right]
\end{align*}
% Non-stationary policies for finite-horizon SSPs can be represented as
% finite-state machines (FSMs) that can be unrolled as tree-structured FSMs.

% Several representations have been developed for efficiently representing
% the MDPs, such as \textit{Relational Dynamic Influence Diagram
%   Language} (RDDL) \cite{sanner10_rddl}, \textit{Probabilistic
%   Planning Domain Definition Language} (PPDDL)
% {\cite{younes04_ppddl}}. These languages separate an MDP domain, which
% consists of parameterized actions, functions, and predicates, from an
% MDP Problem, which expresses the objects, an initial state and a goal
% that needs to be achieved. Without loss of generality, we use PPDDL to
% represent SSPs in this paper.

% \textbf{Example 1.} Fig 2 shows our representation of a pick action
% in the cluttered table domain in a language similar to PPDDL
% (simplified for better readability). The action models the act of
% picking up an object \emph{o} with Gripper \emph{g}, Grasping Pose \emph{gp}, Motion Plan
% \emph{mp}. The precondition specifies that the gripper should be empty,
% grasping pose should be a valid grasping pose for the object, their
% should be a feasible motion plan from the current pose of the robot to
% the grasping pose and the end of the horizon should not have been
% reached. If all of there requirements are met, the robot will pick the
% object with probability 0.9 and drop the object with probability 0.1.

\section{Formal Framework}
\label{sec:formal}

We model stochastic task and motion planning problems as abstracted
SSPs where each action of the SSP (e.g. place) corresponds to an
infinite set of  motion planning problems. The
overall problem is to compute a policy for the SSP along with
``refinements'' that select, for each action in the policy, a specific
motion planning problem and its solution. E.g., the ``high-level''
action for placing a can on a table corresponds to infinitely many
motion planning problems, each defined by a target pose for the
can. The refinement process would thus need to select the pose that
the gripper should be in prior to opening, and a motion plan for each
occurrence of the pickup action in the computed policy.

\noindent\emph{Formalization of abstraction functions}\hspace{10pt} To formalize the necessary abstractions we first introduce some
notation.  We denote states as logical models or structures. We use
the term \emph{logical structures} or \emph{structures} to distinguish
the concept from SDM models. A structure $S$, of vocabulary
$\mathcal{V}$, consists of a universe $\mathcal{U}$, along with a
relation $r^S$ over $\mathcal{U}$ for every relation symbol $r$ in
$\mathcal{V}$ and an element $c^S\in \mathcal{U}$ for every constant
symbol $c$ in $\mathcal{V}$. We denote the value of a term or formula
$\varphi$ in a structure $S$ as $\llbracket
\varphi\rrbracket_S$.%  These values are either True, False, or elements
% of the universe of $S$.
We also extend this notation so that
$\llbracket r \rrbracket_S$ denotes the interpretation of the relation
$r$ in $S$. We consider relations as a special case of
functions.

 We formalize abstractions using first-order
queries \cite{codd72_relational,immerman98_dc} that map structures
over one vocabulary to structures over another vocabulary.  In
general, a first-order query $\alpha$ from $V_\ell$ to $V_h$ defines
functions in $\alpha(S_\ell)$ using interpretations of
$V_\ell$-formulas in $S_\ell$:
$\llbracket r\rrbracket_{\alpha(S_\ell)}(o_1, \ldots o_n)=True$ iff
$\llbracket \varphi^\alpha_r (o_1, \ldots o_n)\rrbracket_{S_\ell}=\emph{True}$, where $\varphi^\alpha_r$ is a
formula over $V_\ell$.

We define \emph{relational abstractions} as first-order queries where
$V_h\subset V_\ell$; the predicates in $V_h$ are defined as identical
to their counterparts in $V_\ell$. Such abstractions reduce the number
of properties being modeled.  Let $\mathcal{U_\ell}$ ($\mathcal{U}_h$)
be the universe of $S_\ell$ ($S_h$) such that $|\U_h | \le
|\U_\ell|$. Function abstractions do not reduce the number of objects
being considered.

Let $\rho: \U_h\rightarrow 2^{\U_\ell}$ be a collection function that
maps elements in $\U_h$ to the collection of $\U_\ell$ elements that
they represent.  E.g.,
$\rho(\emph{Kitchen})=\set{\emph{loc}:\land_i~ \emph{loc} \cdot
  \emph{BoundaryVector}_i<0 }$ where the kitchen has a polygonal
boundary. 

We define an \emph{entity abstraction} $\alpha_\rho$ using the
collection function $\rho$ as
$\llbracket r\rrbracket_{\alpha_\rho(S_\ell)}(\tilde{o}_1, \ldots
\tilde{o}_n)= True$ iff $\exists o_1, \ldots o_n$ such that
$o_i \in \rho(\tilde{o}_i)$ and
$\llbracket \varphi^{\alpha_\rho}_r (o_1, \ldots o_n)\rrbracket_{S_\ell}=\emph{True}$. We omit the subscript $\rho$
when it is clear from context. Entity abstractions define the truth
values of predicates over abstracted entities as the disjunction of the
corresponding concrete predicate instantiations (an object is in the
abstract region ``kitchen'' if it is at any location in that region).
Such abstractions have been used for efficient generalized
planning~\cite{srivastava11_aij} as well as answer set
programming~\cite{zeynep18_aspocp}.

\noindent\emph{STAMP Problems}\hspace{10pt} We define STAMP problems  using
abstractions as follows.

\begin{definition}
  A \emph{stochastic task and motion planning problem (STAMPP)}
  $\langle \M, c_0, \alpha, [\M]\rangle$ is defined using a concrete
  SSP $\M$, its abstraction $[\M]$ obtained using a composition
  of function and entity abstractions, denoted as $\alpha$, and the
  initial concrete configuration of the environment $c_0$.
\end{definition}

Solutions to STAMPPs, like solutions to
an SSP, are policies with actions from the concrete model $\M$.

Let $S$ be the set of abstract states generated when an abstraction
function $\alpha$ is applied on a set of concrete states $X$. For any
$s\in S$, the \emph{concretization function}
$\Gamma_\alpha(s) = \set{x\in X: \alpha(x)=s}$ denotes the set of
concrete states \emph{represented by the abstract state} $s$. For a
set $C\subseteq X$, $[C]_\alpha$ denotes the smallest set of abstract
states representing $C$.  Generating the complete concretization of an
abstract state can be computationally intractable, especially in cases
where the concrete state space is continuous. In such situations, the
concretization operation can be implemented as a \emph{generator} that
incrementally samples elements from an abstract argument's concrete
domain.

% \iftoggle{full}{

\begin{figure}
  \begin{small}
\noindent\emph{\\Place($obj_1$,  $config_1$, $config_2$, $target\_pose$, $traj_1$)}\\
\begin{tabular}{rl}
  \emph{precon} & \emph{RobotAt($config_1$)}, \emph{holding($obj_1$)}, \\ 
                & \emph{IsValidMP($traj_1$, $config_1$, $config_2$)},\\
                & \emph{IsPlacementConfig($obj_1$, $config_2$,} \\
                & \quad \quad \quad \quad \quad \quad \quad \quad \emph{$target\_pose$)}\\
                % & \emph{$\forall obj' \lnot$ Collision($obj'$, $traj_1$)}) \\
  \emph{concrete effect}&  \emph{$\lnot$holding($obj_1$)}, \\
                        & $\forall  traj$ \emph{intersects(vol(obj ,target\_pose)}, \\
                & \emph{sweptVol(robot, traj)$\rightarrow$ Collision($obj_1$, traj)}, \\
                        & \emph{probabilistic:} \\
                        &  \quad 0.8 \big[\emph{RobotAt($config_2$)}, \\
                        & \quad \quad \quad \emph{at($obj_1$, $target\_pose$)}\big]\\
                        & \quad 0.2 \big[\emph{RobotAt($around\_config_2$)}, \\
                        & \quad \quad \quad \emph{at($obj_1$,$around\_target\_pose$})\big] \\
  \emph{abstract effect} & \emph{$\lnot$holding($obj_1$)},\\ 
  & $\forall traj$ \? \emph{Collision($obj_1$, $traj$)}, \\
  &  \emph{probabilistic:} \\
                        &  \quad 0.8 \big[\emph{RobotAt($config_2$)}, \\
                        & \quad \quad \quad \emph{at($obj_1$, $target\_pose$)}\big]\\
                        & \quad 0.2 \big[\emph{RobotAt($around\_config_2$)}, \\
                        & \quad \quad \quad \emph{at($obj_1$,$around\_target\_pose$})\big] \\
\end{tabular}
\end{small}
\caption{\small Concrete (above) and abstract (below) effects of a
  one-handed robot's action for placing an
  object.}\label{fig:place}
\vspace{-20pt}
\end{figure}

\noindent\emph{Example}\hspace{10pt} Consider the specification of a robot's action of placing
an item as a part of an SSP. In practice,
low-level accurate models of such actions may be expressed as
generative models, or simulators. 
Fig. \ref{fig:place} helps identify the nature of abstract
representations needed for expressing such actions. For
readability, we use a convention where preconditions are
comma-separated conjunctive lists and universal quantifiers represent
conjunctions over the quantified variables. Numbers represent the
probability of that outcome.

The concrete, unabstracted, description of this action
(Fig.\,\ref{fig:place}) requires action arguments representing the
object to be placed ($\emph{obj}_1$), the initial and final robot
configurations ($\emph{config}_1$, $\emph{config}_2$), the target pose for the
object ($\emph{target}\_\emph{pose}$), and the motion planning trajectory ($\emph{traj}_1$)
to be used. These arguments represent the choices to be made when
placing an object. The preconditions of \emph{Place} express the
conditions that the robot is in $\emph{config}_1$; it is holding the object
$\emph{obj}_1$; $\emph{traj}_1$ is a motion plan which moves robot from
$\emph{config}_1$ to $\emph{config}_2$ (\emph{IsValidMP)}; $\emph{config}_2$ corresponds to
the object being at the target pose in the gripper
(\emph{IsPlacementConfig}). We ignore the gripper open configuration
for ease in exposition.

This action model specifies two probabilistic effects. The robot moves
to $config_2$ and places the object successfully at $\emph{target}\_\emph{pose}$
with probability 0.8. It moves to some other configuration
$\emph{around}\_\emph{config}_2$ and places the object at a location
$\emph{around}\_\emph{target}\_\emph{pose}$ with probability 0.2. In both cases, the robot
is no longer holding the object and it collides with objects that
lie in the volume swept by the robot while following the trajectory.
The \emph{intersects} predicate is static as it operates on volumes,
while \emph{Collision} changes with the state.

We use entity abstraction to replace each continuous action argument
with a symbol denoting a region that satisfies the precondition
subformulas where that argument occurs. This  may require
Skolemization as developed in prior work~\cite{srivastava14_tmp}.
Effects of abstract actions on symbolic arguments cannot be determined
precisely; their values are assigned by the planning algorithm. E.g.,
it is not possible to determine in the abstract model whether the
placement trajectory will be in collision. Such predicates are
annotated in the set of effects with the symbol \?  (see the abstract effect in
Fig.\,\ref{fig:place}).  This results in a sound abstract
model~\cite{srivastava14_tmp,srivastava16_aaai}.

\section{Algorithmic Framework}
\label{sec:algo}
\subsection{Overall Approach}
\label{sec:algo-overall}
We now describe our approach for computing task and motion policies as
defined above. For clarity, we begin by describing certain choices in
the algorithm as non-deterministic. Variants of our overall approach
can be constructed with different implementations of these choices;
the versions used in our evaluation are described in \ref{sec:optimization}.

Recall that abstract grounded actions $[a]\in[\M]$
(e.g., \emph{Place(cup, config1\_cup, config2\_cup, target\_pose\_cup,
  traj1\_cup)}) have symbolic arguments that can be instantiated to
yield concrete grounded actions $a\in \M$. 

% If the argument
% instantiation satisfies the preconditions of $a$ in a concrete state
% $c$, $\M$ can be used to compute the concrete effects of $a$ on
% $c$. This process requires that it should be possible to evaluate each
% predicate instantiation in a low-level state.

% Of course, doing this evaluation during the search for a plan can be
% prohibitively expensive: one would have to compute all possible
% instantiations of symbolic action arguments and then use $\M$ to
% generate the next possible states. The whole purpose of abstraction is
% to avoid such operations since exploring the space of all possible
% argument instantiations and carrying out action propagation in $\M$
% for each instantiation is computationally intractable, particularly if
% $\M$ is an arbitrary simulator.

Our overall algorithm interleaves computation among the processes of (a)
\emph{concretizing an abstract policy}, (b) \emph{updating the abstraction
 to include predicate valuations for a fixed concretization}, and (c) \emph{computing an abstract
  policy for an updated abstract state}. This is done using the \emph{plan
  refinement graph} (PRG). Every
node \emph{u} in the PRG represents an abstract model $[\M]_u$, an
abstract policy $[\pi]_u$ in the form of a tree whose vertices
represent states and edges represent action applications, 
the current state of the search for concretizations of all actions $a_{j}$
$\in$ $[\pi]_{u}$, and a partial concretization $\sigma_{u}$ for a topological prefix of the policy tree $[\pi]_u$. Each edge \emph{(u,v)} between nodes \emph{u} and
\emph{v} in the PRG is labeled with a partial concretization $\sigma_{u,v}$ and the failed preconditions for the first abstract action in a root-to-leaf path in $[\pi]_u$ which doesn't have a feasible refinement under $\sigma_{u,v}$. Recall that this occurs because the abstract model is lossy and doesn't capture precise action semantics. $[\M]_v$ is the
version of $[\M]_u$ where the predicates corresponding to the failed preconditions  (corresponding to effects with \?, created due to the abstraction discussed in Sec.\,\ref{sec:formal}) have been replaced  with their literal versions that are true under $\sigma_{u,v}$.

ATM-MDP algorithm (Alg.\ref{alg:atm-mdp}) carries out the interleaved search outlined above as follows. It first initializes the PRG with a single node containing an abstract policy for the abstract SSP (line 1). In every iteration of the main loop, it selects a node in the PRG and extracts an unrefined root-to-leaf path from the policy for that node (lines 3-5). It then interleaves the three processes as follows.

\paragraph{a) Concretization of an available policy}
\label{sec:concretization}
Lines 7-13 search for a feasible concretization (refinement) of the partial
path by instantiating its symbolic action arguments with values from their
original non-symbolic domains. Trajectory symbols like $traj_1$ are refined using motion planners. A concretization
$c_0, a_1, c_1, \ldots, a_k, c_k$ of the path
$[s_0], [a_1], [s_1], \ldots, [a_k], [s_k]$ is feasible
starting with a concrete initial state $c_0$ iff
$c_{i+1}\in a_{i+1}(c_{i})$ and $c_{i} \models PRECOND(a_{i+1})$ for $i=0,\ldots,k-1
$. However, it is possible that $[\pi]$ admits no feasible concretization because every
instantiation of the symbolic arguments violates the preconditions of
some action in $\set{\pi_i}$. For example, an infeasible path would have the robot placing a
cup on the table in the concrete state $c_0$, when every possible
motion plan for doing so may be in collision with some object(s).

\paragraph{b) Update abstraction for a fixed concretization}
\label{sec:update}
Lines 16-20 fix a concretization for the partially refined path
selected on line 6, and identify the earliest abstract state in this path whose
subsequent action's concretization is infeasible. This abstract state is updated with the true forms of the violated preconditions that hold in
this concretization, using symbolic arguments. E.g., \emph{Collision($teapot$, $traj\_cup$)}. The rest of the policy after this abstract state is discarded.  A
state update is immediately followed by the computation of a new
  abstract policy (see below).

\paragraph{c) Computation of a new abstract policy}
\label{sec:ssp-invocation}
Lines 21-22 compute a new policy with the updated information computed
under (b). The SSP solver is invoked to compute a new policy from the
updated state; its solution policy is unrolled as a tree of bounded
depth and appended to the partially refined path. This allows the time
horizon of the policy to be increased dynamically.

% \begin{figure}[t!]
%     \centering
%     \includegraphics[width=\columnwidth]{images/batch_pr.png}
%     \caption{\small{ Batch PR node generation: Single PR node generation would only generate one PR child ($B_{1}$); Batch PR node generates $K$ PR children ($B_{1}, B_{2}, ..., B_{K}$) }}
%     \label{fig:batch_pr}
% \end{figure}

Several optimizations can be made while selecting a PRG node to concretize or update in line 3. We used iterative-broadening depth-first search on the PRG  with the max breadth incremented by $5$ in each iteration. 
% \paragraph{PR Node Selection Strategy}
% \label{sec:batch-refinement}

% Every time when complete refinement of the high-level policy is not feasible, generating a single PR child might cause visiting the same PR Node multiple times in the case of dead ends or policies which are difficult to refine.  Instead, multiple PR nodes can be generated from a single PR node by changing the concretization $\sigma$, which would lead to less recurring visits to the same PR node and also allow selection of plans with lesser expected costs. For example, consider the scenario in the Fig \ref{fig:batch_pr}, where there exists multiple high-level policies with different expected costs for different concretizations. If we generate just a single PR node ($B_1$), it might take arbitrarily long depending on the selection of PRG traversal method to come back to the same PR node and generate other high-level policy with lesser expected cost. It might not be even possible to achieve this policy if for any PR node in this subtree rooted at $B_1$ has a complete refinement, which would lead to a suboptimal solutions. Instead, $K$ PR children ($B_{1}, B_{2}, ..., B_{k}$) can be generated at a time to help reduce this recurring visits, and achieve solutions with lesser costs. To do so, we perform the process the process of updating the abstraction and compute the new policy multiple times (line 17) and create PR nodes in batches (Fig. \ref{fig:batch_pr}).

In our implementation the \emph{Compute} variable on line 6 is set to
either \emph{Concretization} or \emph{UpdateAbstraction} with
probability $0.5$. The \emph{explore} parameter on line 9 needs to be set with non-zero probability for a formal guarantee of
completeness, although in our experiments it was set to False.

\begin{algorithm}[t!]
  \begin{small}
    \KwIn{model $[\M]$, domain $\mathcal D$, problem $\mathcal P$, SSP Solver \emph{SSP}, Motion Planner $M$}
    \KwOut{anytime, contingent task and motion policy}
    Initialize PRG with a node with an abstract policy $[\pi]$ for $P$
    computed by \emph{SSP}\;
    \While{solution of desired quality not found}
    {
      PRNode $\gets$ GetPRNode()\;
      $[\pi]$ $\gets$ GetAbstractPolicy($[\M]$, PRNode, {
        $\mathcal D$}, {$\mathcal P$}, \emph{SSP})\;
      path\_to\_refine $\gets$ GetUnRefinedPath($[\pi]$)\;
      Compute $\gets$
      NDChoice\{\emph{Concretization}, \emph{UpdateAbstraction}\}\;
      \If{Compute = \emph{Concretization}}{
        \While{$[\pi]$ has an unrefined path and resource limit is not
          reached}{
          \If{explore\tcp{non-deterministic}}{
            replace a suffix of partial\_path with a random action\;
          }
          search for a feasible concretization of path\_to\_refine\;
        }
      }
      \If{Compute = \emph{UpdateAbstraction}}{
        partial\_path $\gets$ GetUnrefinedSuffix(PRNode, path\_to\_refine)\;
        $\sigma \gets$ ConcretizeLastUnrefinedAction($[\pi]$)\;
        failure\_reason $\gets$ GetFailedPrecondition($[\pi]$, $\sigma$
        )\;
        updated\_state $\gets$ UpdateState($[\pi]$, failure\_reason)\;
        % random\_action $\gets$ choose\_random\_action(${\cal D}$)\;
        % next\_state $\gets$ apply\_action(random\_action)\;
        $[\pi']\gets$ merge($[\pi]$, solve(updated\_state, $G$, $[\M]$))\;
        generate\_new\_pr\_node($[\pi']$, $[\M]$)\;
      }
    }
  \end{small}
  \caption{\small ATM-MDP Algorithm}\label{alg:atm-mdp}
\end{algorithm}

\subsection{Optimizations and Formal Results}
\label{sec:optimization}
We develop the basic algorithm outlined above (Alg.\,\ref{alg:atm-mdp}) along two major directions: we enhance it to facilitate anytime computation and to improve the search for concretizations of abstract policies.

\subsubsection{Anytime computation for task and motion policies}
The main computational challenge for the algorithm is that the number
of root-to-leaf (RTL) branches grows exponentially with the time
horizon and the contingencies in the domain. Each RTL branch has a certain
probability of being encountered; refining it incurs a computational
cost. Waiting for a complete refinement of the policy tree results in wasting a lot of
time as most of the situations have a very low probability of being encountered.  The optimal selection of the paths to refine within a fixed
computational budget can be reduced to the knapsack
problem. Unfortunately, we do not know the precise
computational costs required to refine a path. However, we can approximate this cost depending on the number of actions and the size of the domain of the arguments in those actions. Furthermore, the knapsack problem is NP-hard. However, we can compute provably good approximate solutions to this problem using a greedy approach: we prioritize the selection of a path to refine based on the probability
of encountering that path \emph{p} and the estimated cost of
refining that path \emph{c}. We compute $p/c$ ratio for all the paths and select the unrefined path with largest ratio for refinement. 

\begin{figure}[t!]
    \centering
    \includegraphics[height=1in]{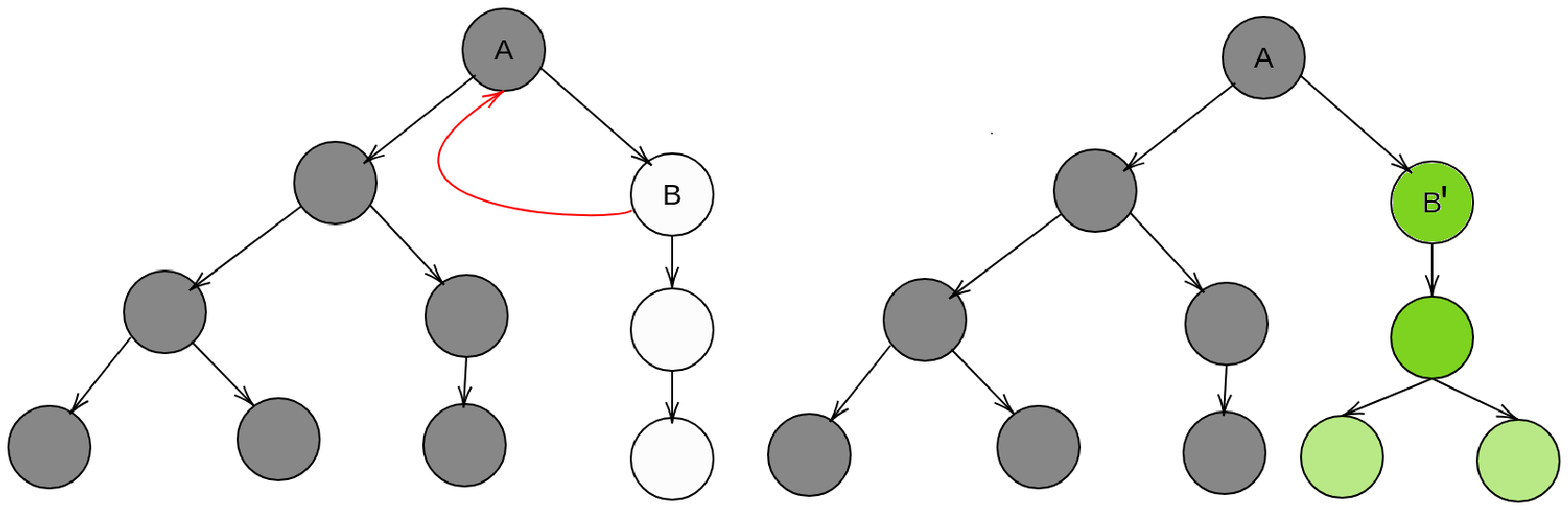}
    \caption{Left: Backtracking from node B invalidates the refinement of subtree rooted at A. Right: Replanning from node B.}
    \label{fig:ptree}
    \vspace{-10pt}
\end{figure}

\subsubsection{Search for concretizations}
Sample-based backtracking search for concretization of symbolic
variables~\cite{srivastava14_tmp} suffers from a few limitations in
stochastic settings that are not present in deterministic
settings. Fig.\,\ref{fig:ptree} illustrates the problem. In this
figure, grey nodes represent actions in the policy tree that have
already been refined; the refinement for \emph{B} is being
computed. White nodes represent the nodes that still require
refinement. If backtracking search changes the concretization for B's
parent (Fig.\,\ref{fig:ptree}, left) it will invalidate the
refinements made for the entire subtree rooted at that node. Instead,
it may be better to compute an entirely new policy for B (effectively
jumping to the UpdateAbstraction mode of computation on line 16 from
line 13).

 Thm.\,\ref{thm:knapsack} formalizes the
anytime performance of ATM-MDP and Thm.\,\ref{thm:pc}
shows that our solution to this problem is probabilistically complete.
Additional details about these results are available in
the extended version of the
paper~\cite{tmp_full_version}.

\begin{theorem} \label{thm:knapsack} Let $t$ be the time since the start of the algorithm
  at which the refinement of any $RTL$ path is completed. If
  path costs are accurate and constant then the total probability of
  unrefined paths at time $t$ is at most $1 - opt(t)/2$, where
  $opt(t)$ is the best possible refinement (in terms of the
  probability of outcomes covered) that could have been achieved in
  time $t$.
\end{theorem}
\iftoggle{full}
{
    \begin{proof}
   (\emph{Sketch})
    The proof follows from the fact that the greedy algorithm achieves a
    2-approximation for the knapsack problem. In practice, we estimate the
    cost as $\hat{c}$, the product of measures of the true domains of each
    the symbolic argument in the given RTL. Since, $\hat{c}\ge c$ modulo
    constant factors, the priority queue never can only underestimate the
    relative value of refining a path, and the algorithm's coverage of
    high-probability contingencies will be closer to optimal than the
    bound suggested in the theorem above. This optimization gives a user
    the option of starting execution when a desired value of the probability
    of covered contingencies has been reached. 
    \end{proof}
}{

  \begin{proof} (\emph{Sketch}) Let $c$ be the cost of refining some RTL path and $\hat{c}$ be an
  approximation of it. The proof follows from the fact that the greedy
  algorithm achieves a 2-approximation for the knapsack problem and
  that for all RTL paths, $\hat{c} \ge c$. So the priority queue will
  never underestimate the relative costs, and algorithm's coverage of
  high-probability contingencies will be no further from the optimal
  than the bound suggested in the theorem.  \end{proof}}
% Numerous additional optimizations could be used to further improve the
% performance of this approach in future work. In particular, better
% strategies and/or statistical learning could be used in place of the
% probabilistic choices in the search for concretizations and in the
% selection of the mode of computation (line 7).

\begin{theorem}
  \label{thm:pc}
  If there exists a proper policy that reaches the goal within
  horizon \emph{h} with probability \emph{p}, and has feasible
  low-level refinement, then Alg.\,\ref{alg:atm-mdp} will find it with
  probability $1.0$ in the limit of infinite samples.\end{theorem}

% \begin{comment}
%   \textbf{Proof:} Let the length of longest prefix common with a
%   solution that has a possible refinement as motion plan, be defined
%   as a \emph{measure of correctness}. Assume that at some point of
%   time we get a plan with measure of correctness \emph{k} and try to
%   refine $k+1^{th}$ action which does not have valid refinement. In
%   this case, the algorithm selects a random action and then generates
%   a plan to reach the goal. As the number of actions is finite, this
%   ensures that at some point of time, the algorithm will have plan
%   with measure of correctness \emph{k+1}. If the solution plan has
%   length \emph{l}, once algorithm finds a plan with measure of
%   correctness \emph{l}, it is guaranteed to find the refinement by
%   changing the concretization of the actions. In practice,
%   communicating reason of infeasibility made it unnecessary to try
%   random actions in all of our tests.
% \end{comment}
\iftoggle{full}{
\begin{proof} (\emph{Sketch}) Let $\pi_p$ be a proper policy. Consider a policy
  $\pi$ in the PRG; let $k$ denote the minimum depth up to which
  $\pi_p$ and $\pi$ match. $k$ will be used as a \emph{measure of
    correctness}. When $\pi$'s PRG node is selected, suppose we try to
  refine one of the child nodes of depth $k+1$ in the partial path
  that had the $k$-length prefix consistent with the solution.  The
  algorithm selects the correct child action with non-zero probability
  under the \emph{explore} steps (line 11), and then generates a plan
  to reach the goal from the resultant state. The finite number of
  discrete actions and the fixed horizon ensures that at in time
  bounded in expectation, ATM-MDP will generate a policy with the  measure
  of correctness $k+1$. Once the algorithm finds the policy with
  the measure of correctness \emph{h}, it stores it in the PRG and is
  guaranteed to find feasible refinements with probability one if the
  measure of these refinements under the probability-density of the
  generators is non-zero.
\end{proof}
}{
  \begin{proof} (\emph{Sketch}) Let $\pi_p$ be a proper policy. For
    some policy $\pi$ in PRG, let $k$ denote the minimum depth up to
    which $\pi_p$ and $\pi$ match. If there are no feasible
    refinements possible for an action at depth $k+1$ in $\pi$, then
    the \emph{explore} step (line 11) would replace that action such that
    it matches the action at depth $k+1$ in $\pi_p$ with some non-zero
    probability (given that actions are finite). Once the algorithm
    finds policy $\pi$ which matches $\pi_p$, the backtracking search
    will find a feasible refinement if the measure of these
    refinements under the probability density of  generators is
    non-zero.
  \end{proof}
}

\section{Empirical Evaluation}
\label{sec:results}
We implemented the presented framework using an open-source
implementation from MDP-Lib github repository ~\cite{mdplib} of LAO*~\cite{hansen01_lao} as the SSP solver, the
OpenRAVE~\cite{diankov10_openrave} robot simulation system along with its collision checkers, CBiRRT implementation from PrPy suite ~\cite{prpy} for
motion planning. Since there are no common benchmarks for evaluating
stochastic task and motion planning problems, we evaluated our
algorithm on 7 diverse and challenging test problems over 4 domains and evaluated 5 of those problems with physical robot systems. In practice, fixing the horizon $h$ a priori can render some problems unsolvable. Instead, we implemented a variant that dynamically increases the horizon until the goal is reached with probability greater than $0$. 
We evaluated our approach on a variety of problems where combined task and motion planning is necessary. The source code and the videos for our experiments experiment can be found at \href{https://aair-lab.github.io/atam_full.html}{\url{https://aair-lab.github.io/stamp.html}}. 

% \begin{figure}[t!]
%     \centering
%     \includegraphics[width=\columnwidth]{images/cans_hist.eps}
%     \caption{\small Time taken to compute  abstract policies with complete
%       motion planning refinements for randomly generated problems (left: cluttered domain with 15 cans, center: cluttered domain with 20 cans, right: cluttered domain with 25 cans).}
%     \label{fig:results1}
% \end{figure}

\begin{figure}
  \centering
  \vspace{1em}
 \footnotesize
 \begin{tabular}{|l|c|c|}
        \hline
        Problem & \% Solved & Avg. Time (s)  \\ \hline
         Cluttered-15 & 95 & 1093.71 \\
         Cluttered-20 & 79 & 1144.85 \\
         Cluttered-25 & 74 & 1392.83 \\
         Aircraft Inspection & 100 & 1457.08 \\
         $3\pi$ & 100 & 1312.83 \\
         Tower-12 & 100 & 1899.73 \\
         Twisted-Tower-12 & 98 & 1984.29 \\ 
         Domino $(n = 10, k = 2)$ & 100 & 98.64 \\ 
         Domino $(n = 10, k = 3)$ & 100 & 350.63 \\ 
         Domino $(n = 15, k = 2)$ & 100 & 179.60 \\ 
         Domino $(n = 15, k = 3)$ & 100 & 631.91 \\ 
         Domino $(n = 20, k = 2)$ & 100 & 350.60 \\ 
    Domino $(n = 20, k = 3)$ & 100 & 590.60 \\ \hline
    \end{tabular}
    \caption{Summary of times taken to solve the test problems. Timeout for cluttered table, aircraft inspection, and Domino: 2400 seconds, building Keva stuctures: 4000 seconds.}
    \vspace{-3em}
    \label{fig:time_result}
\end{figure}

\begin{figure}[t!]
\end{figure}

% \begin{figure}[t!]
%     \centering
%     \includegraphics[width=\columnwidth]{images/collage_keva.png}
%     \caption{\small Structures built using Keva planks. Left: spiral tower; center: square tower; right: stacked $\pi$.  }
%     \label{fig:keva_strucutres}
% \end{figure}

\begin{figure*}[t!]
  \begin{subfigure}{0.55\textwidth}
    \centering
    \vspace{1em}
      \includegraphics[height=1in]{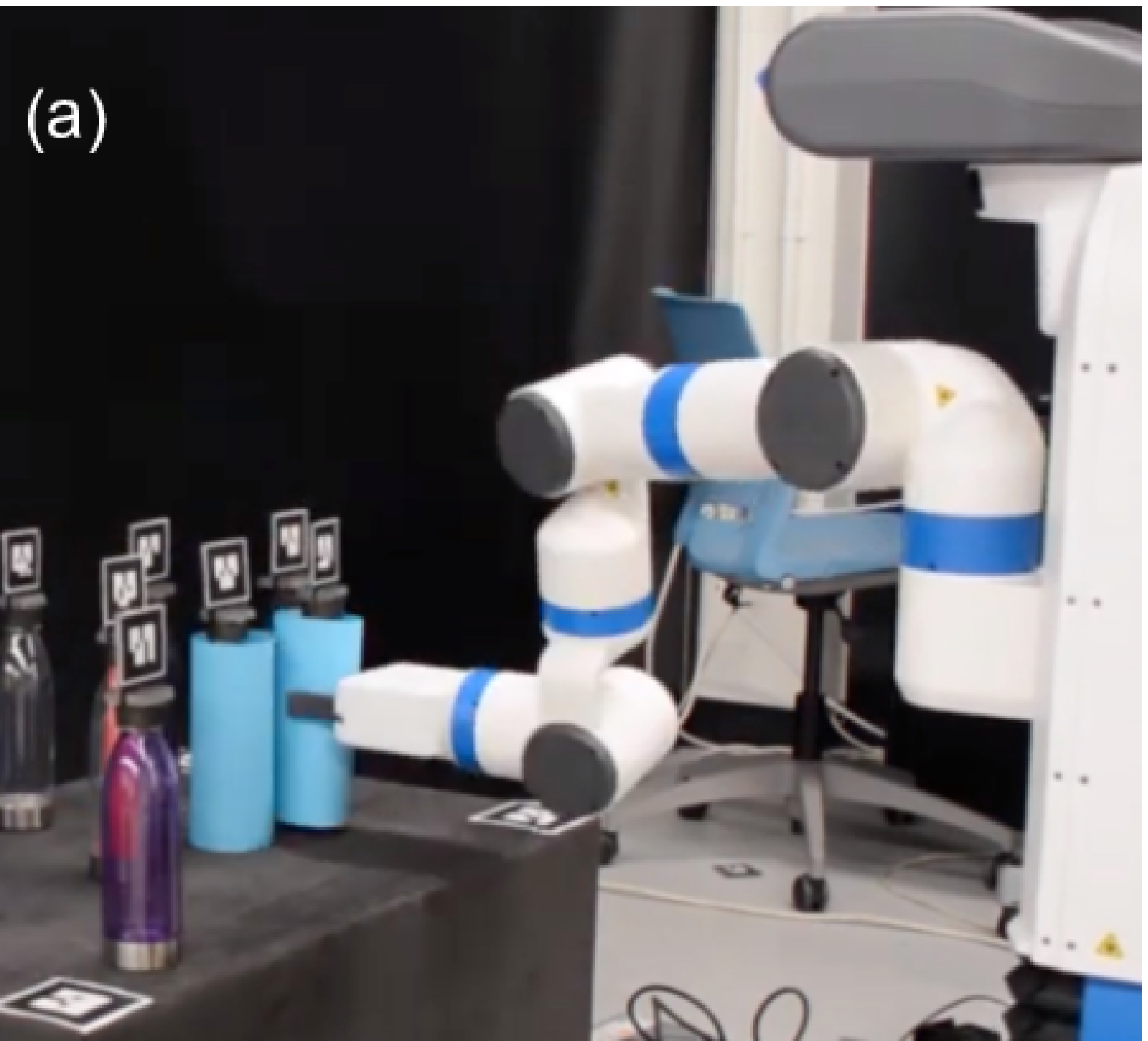}
      \includegraphics[height=1in]{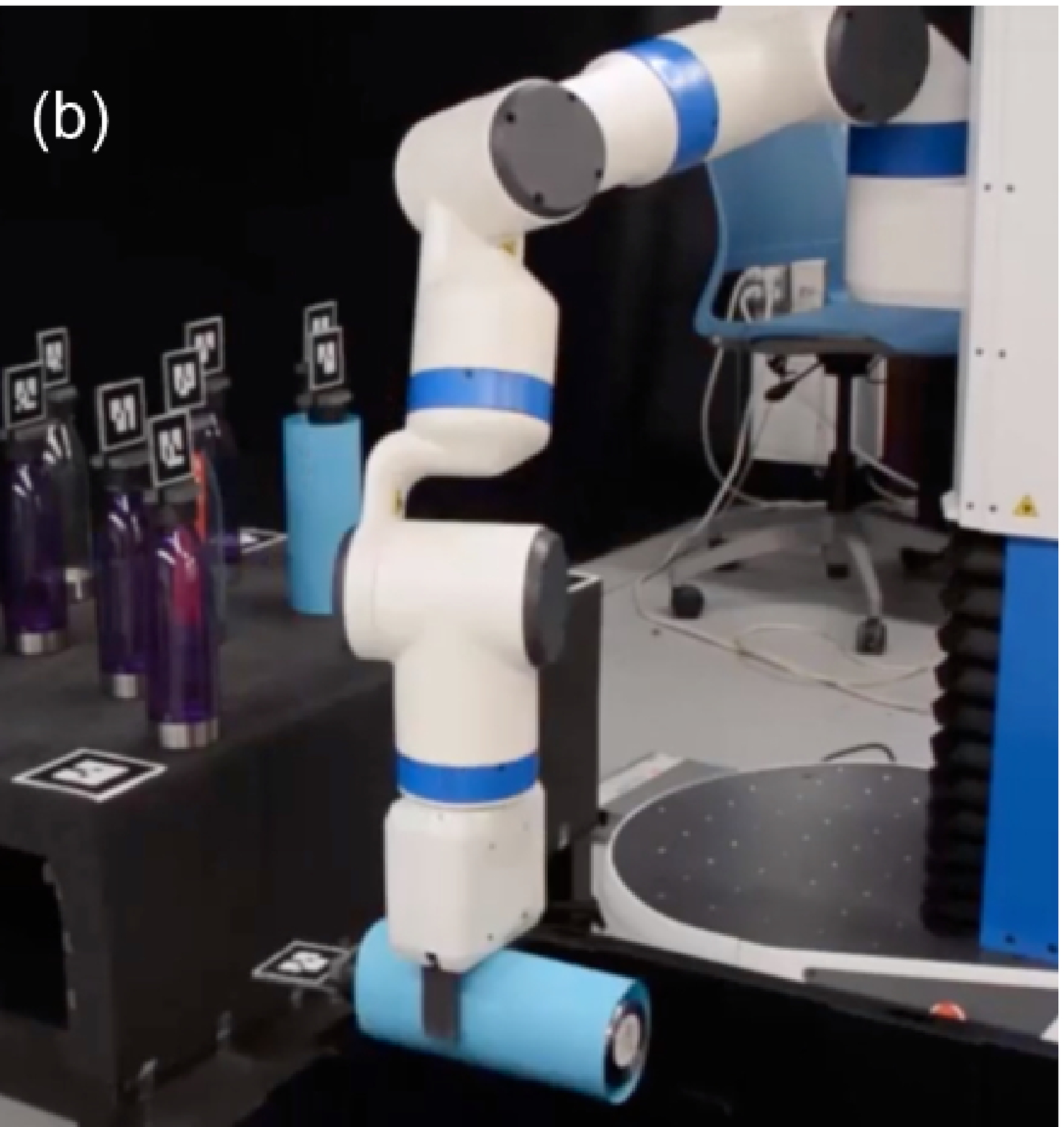}
      \includegraphics[height=1in]{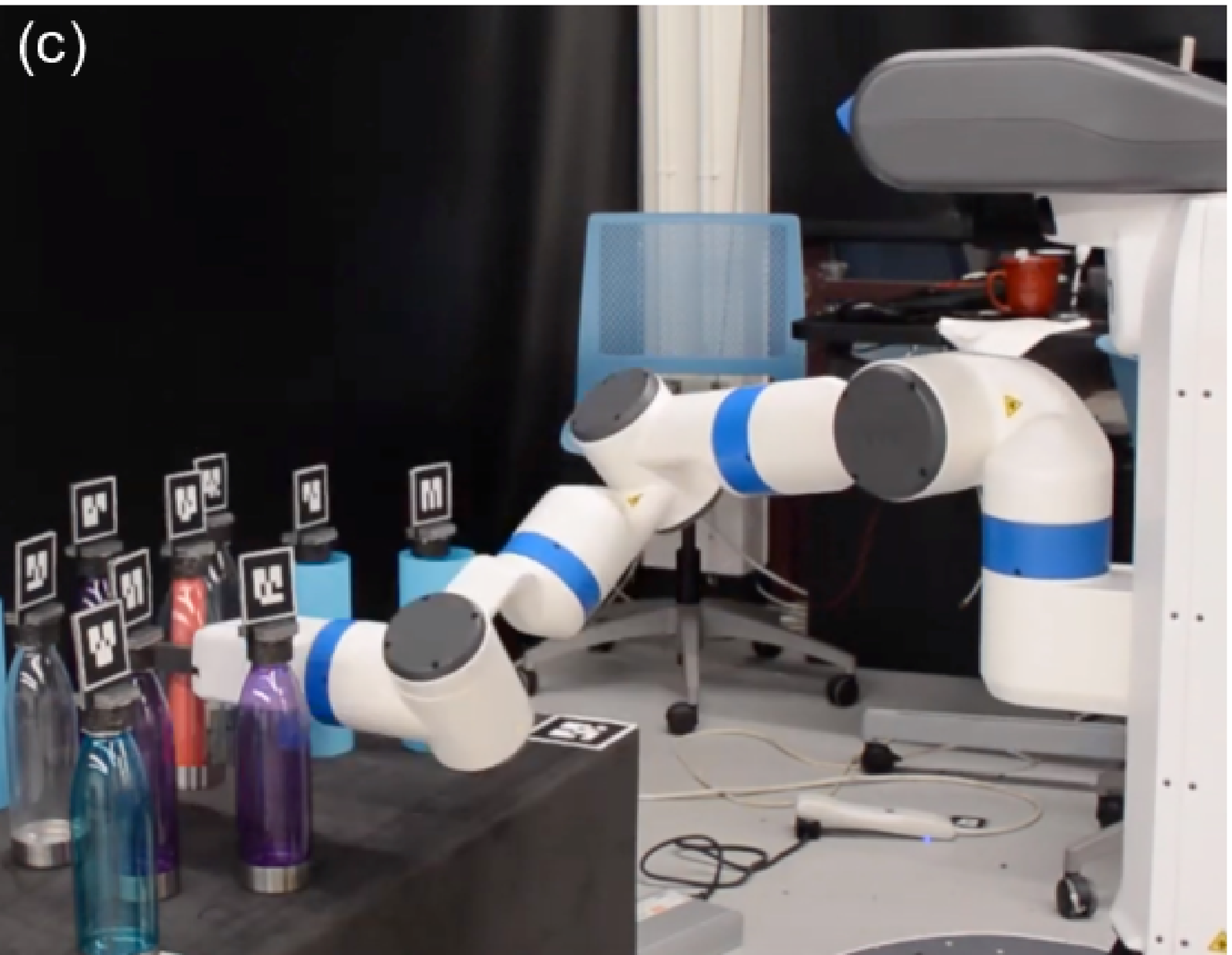}
      \caption{\footnotesize The Fetch mobile manipulator uses a STAMP policy to
        pickup a target bottle while avoiding those that are likely to
        be crushed. It replaces a bottle that wasn't crushed (left), 
        discards a bottle that was crushed (center) and picks up the target
        bottle (right).  }
    \end{subfigure}
    \hspace{0.5cm}
    \begin{subfigure}{0.37\textwidth}
      \centering
      \vspace{1em}
    \includegraphics[height=1in]{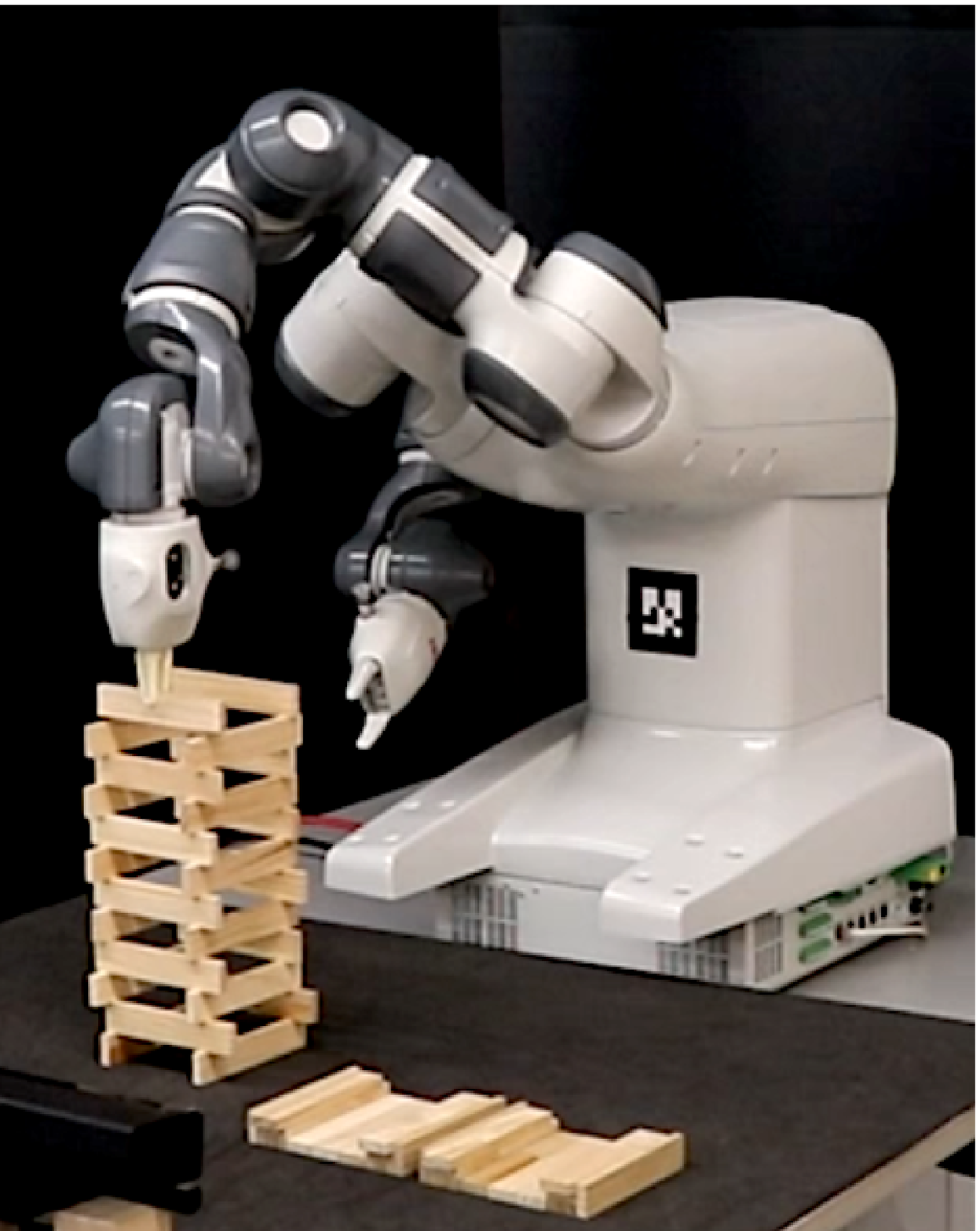}
    \includegraphics[height=1in]{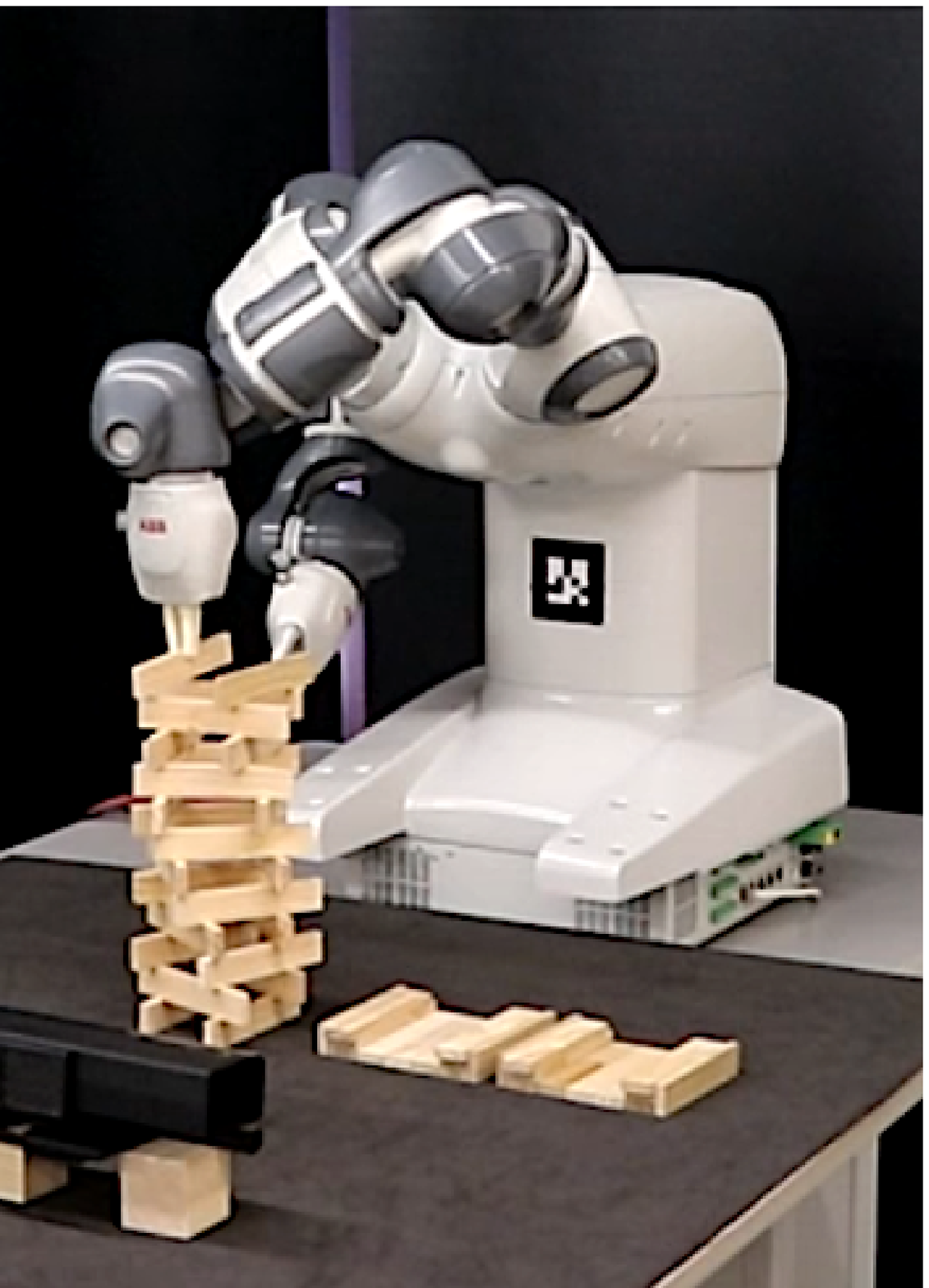}
    \includegraphics[height=1in]{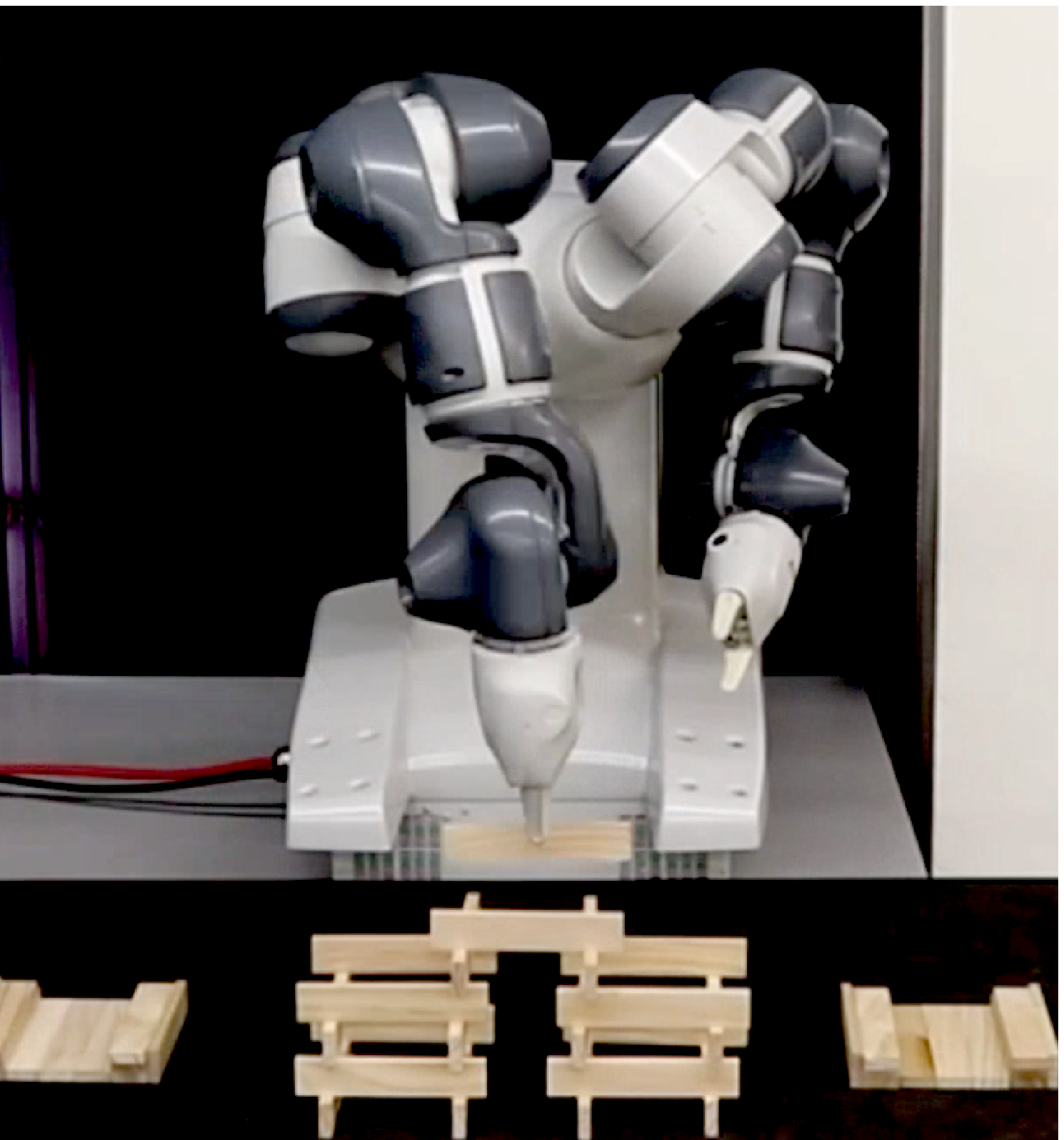}
    \caption{\footnotesize ABB YuMi builds Keva structures using a STAMP policy:
        12-level tower (left), twisted 12-level tower (center), and
        $3$-towers (right).}
    \end{subfigure}
    \caption{Photos from our evaluation using the Fetch and YuMi
      robots. Videos are available at
      \href{https://aair-lab.github.io/stamp.html}{\url{https://aair-lab.github.io/stamp.html}}. }
    \label{fig:keva_strucutres}
    \label{fig:fetch_exp}
    \vspace{-10pt}
\end{figure*}

\begin{figure}[t!]
    \centering
   \includegraphics[width=\columnwidth]{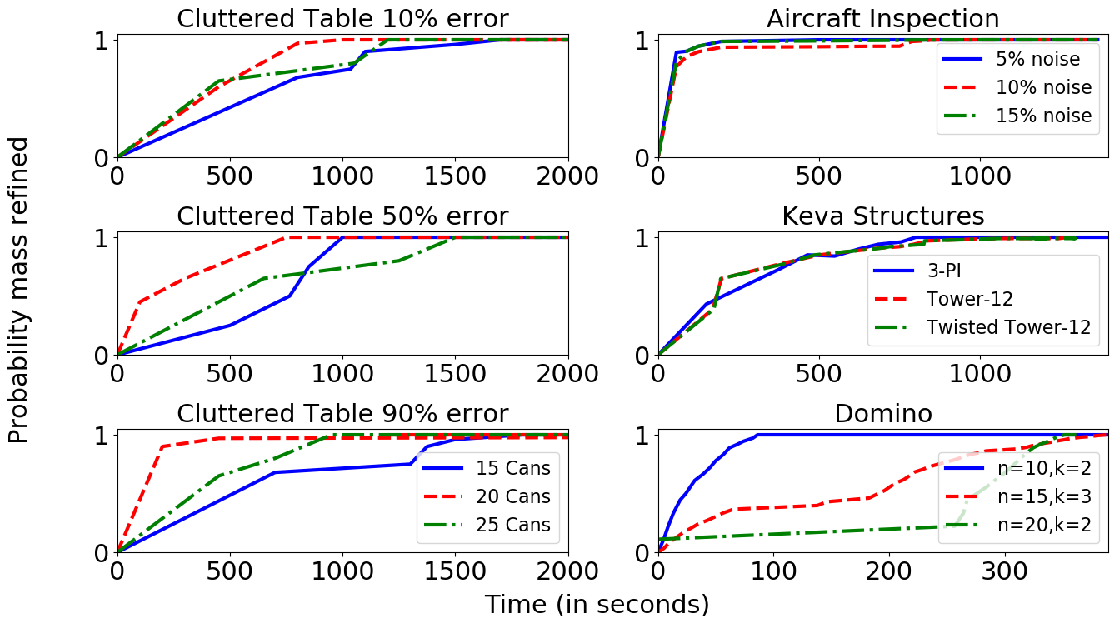}
    \caption{Anytime performance of ATM-MDP, showing the time in seconds (x-axis) vs. probability mass refined (y-axis). } \vspace{-1em}
    \label{fig:anytime_result}
\end{figure}

% \subsubsection{Cluttered Table} In this problem the Fetch robot needs
% to pick up a specific object from a cluttered table. The target object
% can be obstructed by different objects which need to picked and
% placed at different locations to reach the final object. The actions
% available to the robot are to pick up an object, place an object, and
% to move around the table. Generators for the concretization of the
% actions include generating the grasping poses, put-down poses and
% target base poses. The action of picking up an object succeeds with
% probability 0.8; the object falls back onto the table with probability
% 0.2. We increase the number of cans on the table to increase the
% complexity of this problem. Fig.\,\ref{fig:results1} shows the results
% for the time taken to solve 100 randomly generated instances for three configurations of the environment with 15, 20, and 25 number of cans with horizon initially kept to 6.

\paragraph{Cluttered Table} In this problem, we have a table cluttered with cans having different probabilities of being crushed when grabbed by the robot. Some cans are delicate and are highly likely to be crushed  when the robot grabs them, incurring a high cost (probability for crushing was set to 0.1, 0.5 \& 0.9 in different experiments in Fig. \ref{fig:anytime_result}(a)), while others are normal and are less likely to be crushed (with probability set to 0.05). The goal of the robot is to pick up a specific can. We used different numbers of cans (15, 20, 25), and different random configurations of cans to extensively evaluate the proposed framework. We also used this scenario to evaluate our approach in the real-world (Fig. \ref{fig:fetch_exp}(a)) using the Fetch mobile manipulation robot~\cite{wise16_fetch}.
  
% \subsection{Aircraft Inspection} In this problem, a UAV needs to
% inspect an airplane. The available actions are to fly to a certain
% region, go to the charging station and charge, and to inspect a
% certain component. While trying to fly from one location to other
% location, the UAV may drift to a different region with probability
% 0.05. Each operation consumes a certain amount of battery, but this
% cannot be computed at the high-level since the high-level abstraction
% cannot reason with trajectories. There is a battery recharge station
% which can be reached from anywhere in the environment on reserve
% power. Generators for this problem include sampling target locations
% for move actions as well as the waypoints used to envelope a component
% for the inspection action. The algorithm needs to come up with the
% sequence of actions to examine the required parts with valid non-colliding trajectories while keeping sufficient battery at each time
% step. Fig.\,\ref{fig:results2} shows the results for probability
% reaching the goal refined with the percentage of nodes in policy tree
% refined. Empirical evaluations show that it takes less than 1\% of nodes refined and less than 100 seconds to  whereas entire policy tree refinement takes more than 4600 seconds.

\paragraph{Aircraft Inspection}
In this problem, an unmanned aerial vehicle (UAV) needs to inspect possibly faulty parts of an aircraft. Its goal is to locate the fault and notify the supervisor about it. However, its sensors are not accurate and may fail to locate the fault with some non-zero probability (failure probability was set to 0.05, 0.1, \& 0.15 for experiments in Fig. \ref{fig:anytime_result}(b)) while inspecting the location; it may also drift to another location while flying. Charging stations are available for the UAV to dock and charge itself. All movements use some amount of battery charge depending on the length of the trajectory, but the high-level planner cannot determine whether the current level of the  battery is sufficient for an action as it doesn't have access to precise trajectories.  This makes it necessary for the high-level to obtain feedback from the low-level to solve the problem.

\paragraph{Domino} In this problem, the YuMi robot~\cite{yumi} needs to pick up a domino from a table that has $n$ dominos on it. It has to notify the human about toppled dominos. While trying to pick up a domino, $k$ domino's on each side can topple adding up to $2^{2k}$ contingencies that might need refinement. 

\paragraph{Building structures using Keva planks}
In this problem, the YuMi robot~\cite{yumi} needs to build different structures using Keva planks. Keva planks are laser cut wooden planks with uniform geometry. Fig.\,\ref{fig:keva_strucutres}(b) and Fig.\,\ref{fig:domainFig} show some of the target structures.  Planks are placed one at a time by a user after each pickup and placement by the YuMi. Each new plank may be placed at one of a few predefined locations, which adds uncertainty in the planks' initial location. For our experiments, two predefined locations were used to place the planks with a probability of 0.8 for the first location and a probability of 0.2 for the second location. In this problem, hand-written goal conditions are used to specify the desired target structure. The YuMi~\cite{yumi} needs to create a task and motion  policy for successively picking up and placing planks to build the structure. There are infinitely many configurations in which one plank can be placed on another, but the abstract model blurs out different regions on the plank. The put-down pose generator uses the target structure to concretize each plank's target put-down pose. State-of-the-art SSP solvers fail to compute abstract solution policies for structures of height greater than 3 for this problem. However, these structure-building problems exhibit repeating substructure every 1-2 layers that reuse minor variants of the same abstract policy. We used this observation to develop an SSP solver that incrementally calls LAO* to compute  iterative policies. The results for Keva structures use this solver. In addition to the test problems shown in Fig.\,\ref{fig:time_result} this allows our approach to scale to more complex problems such as 3-towers (Fig.\,\ref{fig:keva_strucutres}). Approaches for generalized planning~\cite{srivastava2008AAAI,bonet2009,hu2011,srivastava11_aij} could be used to automatically extract and utilize such patterns in other problems with repeating structures.

\subsection{Analysis of Results} Fig.\,\ref{fig:anytime_result} shows the anytime characteristics of our approach in all of the test domains. The y-axis shows the probability with which the policy available at any time during the algorithm's computation will be  able to handle all possible execution-time outcomes, and the x-axis shows the time (seconds) required to refine that probability mass.  

These results indicate that in all of our test domains, the refined
probability mass increases rapidly with time so that about 80\% of
probable executions are covered within about 30\% of the computation
time. Fig.\,\ref{fig:anytime_result} also shows that refining the
entire policy tree requires a significant time. This reinforces the
need for an anytime solution in such problems.

Fig.\,\ref{fig:time_result} shows average times taken to compute complete STAMP 
policies for our test problems. These values are
averages of 50 runs for  cluttered table, 20 runs for  aircraft
inspection and 15 runs for  Keva structures.  Aircraft inspection
problems and Keva structure problems required fewer runs because their
runtimes showed negligible variance.

% Fig.\,\ref{fig:time_result} shows how using $p/c$ ratio to select RTL paths for refinement reduces the number of unanticipated action while executing the policy. Fig.\,\ref{fig:time_result} shows the number of total actions executed and the number of unanticipated actions for the Keva domain. 

\begin{comment}

  Figure \ref{fig:results1} shows the time taken to solve the entire problem
  over 100 randomly generated instances for domain 1 with Twisted15, 20, and 25 cans respectively from left. This time includes computing abstract policy, generating tree-structured policy, evaluating and
  refining all the nodes in the policy.

\end{comment}

\section{Other Related Work}
\label{sec:related}
There has been a renewed interest in integrated task and motion
planning algorithms. Most research in this direction has been focused
on deterministic
environments \citep{cambon09_asymov,plaku10_sampling,dornhege12_semantic, kaelbling11_hierarchical,garrett15_ffrob,dantam16_incremental,garrett18_sampling}. Kaelbling
and Lozano-Perez \cite{kaelbling13_hpnPOMDP} consider a partially
observable formulation of the problem. Their approach utilizes
regression modules on belief fluents to develop a regression-based
solution algorithm. While they address the more general class of
partially observable problems, their approach follows a process of
online, incremental discretization and does not address the
computation of branching policies, which is the focus of this
paper. Sucan and Kavraki \cite{sucan12_tmp_mdp} use an explicit
multigraph to represent the problem for which motion planning
refinements are desired. Other approaches~\cite{hadfield15_modular}
address problems where the high-level formulation is deterministic and
the low-level is determinized using most likely observations. Our
approach uses a compact, relational representation; it employs
abstraction to bridge SSP solvers and motion planners and solves the
overall problem in anytime fashion. Preliminary versions of this work~\cite{srivastava2018anytime,Shah2019AnytimeIT} were presented at non-archival venues and did not include the full formalization and optimizations required to solve the realistic tasks prsented in this paper.

Several approaches utilize abstraction for solving MDPs~\cite{hostetler14_state,bai16_markovian,li06_abstractMDP,singh95_abstractRL}. However,
these approaches assume that the full, unabstracted MDP can be
efficiently expressed as a discrete MDP. Marecki et
al. \cite{marecki06_cmdp} consider continuous-time MDPs with finite
sets of states and actions. In contrast, our focus is on MDPs with
high-dimensional, uncountable state and action spaces. Recent work on
deep reinforcement learning (e.g.,
\citep{hausknecht16_iclr,mnih15_drl}) presents approaches for using
deep neural networks in conjunction with reinforcement learning to
solve short-horizon MDPs with continuous state spaces. These
approaches can be used as primitives in a complementary fashion with
task and motion planning algorithms, as illustrated in recent
promising work by Wang et al. \cite{wang18_active}.

%Efforts towards improved representation languages are orthogonal to
% our contributions~\citep{fox02_pddl+}. The fundamental computational
% complexity results indicating growth in complexity with increasing
% sizes of state spaces, branching factors, and time horizons remain
% true regardless of the solution approach taken. It is unlikely that a
% uniformly precise model, a simulator at the level of precision of
% individual atoms, or even circuit diagrams of every component used by
% the agent will help it solve the kind of complex tasks on which humans
% would appreciate the assistance. On the other hand, not using any model at
% all would result in dangerous agents that would not be able to safely
% evaluate the possible outcomes of their actions. Our results show that
% these divides can be bridged using hierarchical modeling and solution
% approaches that simplify the representational requirements and offer
% computational advantages that could make autonomous robots feasible in
% the real world.

%%% Local Variables:
%%% mode: latex
%%% TeX-master: "main"
%%% End:
\section*{ACKNOWLEDGMENTS}
We thank Nishant Desai, Richard Freedman, and Midhun Pookkottil Madhusoodanan for their help with an initial implementation of the presented work. This work was supported in part by the NSF under grants IIS 1844325, IIS 1909370, and OIA 1936997.

% \section{References}
\bibliographystyle{IEEEtran}

\bibliography{icaps}

\end{document}